\documentclass{article}

\usepackage[preprint]{neurips_2021}

\usepackage[utf8]{inputenc} 
\usepackage[T1]{fontenc}    
\usepackage{hyperref}       
\usepackage{url}            
\usepackage{booktabs}       
\usepackage{amsfonts}       
\usepackage{nicefrac}       
\usepackage{microtype}      
\usepackage{xcolor}         
\usepackage[utf8]{inputenc} 
\usepackage[T1]{fontenc}    
\usepackage{hyperref}       
\usepackage{url}            
\usepackage{booktabs}       
\usepackage{amsfonts}       
\usepackage{nicefrac}       
\usepackage{microtype}      
\usepackage{xcolor}         
\usepackage{natbib}
\usepackage{hyperref}
\usepackage{url}
\usepackage{graphicx}
\setcitestyle{round}
\usepackage{xcolor}
\usepackage{wrapfig}
\usepackage{caption}
\usepackage{float}
\usepackage[position=bottom]{subfig}
\usepackage{balance}

\usepackage{amsmath,amsfonts,bm}

\def\eqref#1{equation~\ref{#1}}

\def\1{\bm{1}}

\DeclareMathAlphabet{\mathsfit}{\encodingdefault}{\sfdefault}{m}{sl}
\SetMathAlphabet{\mathsfit}{bold}{\encodingdefault}{\sfdefault}{bx}{n}

\usepackage{multirow}
\usepackage{times}  
\usepackage{helvet}  
\usepackage{courier}  

\title{Conditional Synthetic Data Generation for Personal Thermal Comfort Models}

\author{%
  Hari Prasanna Das and Costas J. Spanos \\
  Department of Electrical Engineering and Computer Sciences\\
  University of California\\
  Berkeley, CA, 94720 \\
  \texttt{\{hpdas,spanos\}@berkeley.edu} \\
}

\begin{document}

\maketitle

\begin{abstract}
    Personal thermal comfort models aim to predict an individual's thermal comfort response, instead of the average response of a large group. Recently, machine learning algorithms have proven to be having enormous potential as a candidate for personal thermal comfort models. But, often within the normal settings of a building, personal thermal comfort data obtained via experiments are heavily class-imbalanced. There are a disproportionately high number of data samples for the \textquotedblleft Prefer No Change\textquotedblright class, as compared with the \textquotedblleft Prefer Warmer\textquotedblright and \textquotedblleft Prefer Cooler\textquotedblright classes. Machine learning algorithms trained on such class-imbalanced data perform sub-optimally when deployed in the real world. To develop robust machine learning-based applications using the above class-imbalanced data, as well as for privacy-preserving data sharing, we propose to implement a state-of-the-art conditional synthetic data generator to generate synthetic data corresponding to the low-frequency classes. Via experiments, we show that the synthetic data generated has a distribution that mimics the real data distribution. The proposed method can be extended for use by other smart building datasets/use-cases.
\end{abstract}
\section{Introduction}
Humans spend more than 90\% of their day indoors, where their well-being, performance and energy consumption are demonstrably linked to thermal comfort. But, study shows that only 40\% of commercial building occupants are satisfied with their thermal environment~\cite{graham2021lessons}. There has been significant amount of research done to develop models to accurately predict thermal comfort metrics for occupants in a building. Contrary to conventional group-based thermal comfort models, personal thermal comfort models~\cite{liu2018personal} focus on developing thermal comfort predictors at a building occupant level. They have proved efficient in human-centric cyber-physical systems to efficiently regulate the building control systems, as well as to understand the correlation between human factors affecting comfort. The general process is to conduct experiments with human subjects and collect their physiological signals along with other environmental parameters, and thermal sensations and preference. Then prediction models are trained to predict the thermal preference that governs the thermal comfort management actuators/ controllers. Recently, machine learning models have been introduced to successfully predict thermal comfort.

In real life, often the thermal comfort data obtained is highly class-imbalanced. For instance, in the experiment in ~\citet{liu2019personal}, on an average for each subject, around 65\% of the data belonged to the \textquotedblleft Prefer No Change\textquotedblright class, and the rest equally divided between the \textquotedblleft Prefer Warmer\textquotedblright and \textquotedblleft Prefer Cooler\textquotedblright classes. Machine learning algorithms require high amounts of varied data for efficient performance. Under such class-imbalance, machine learning algorithms perform sub-optimally. In case of buildings, having access to significant amounts of real data for the low-frequency classes, with human subjects is hard and expensive. To balance the classes, recent works have proposed undersampling the high-frequency class to match the count with low-frequency classes, or oversampling the low-frequency classes to match with the high-frequency class. In the former method, there is loss of information, which is undesirable, and in the latter case, there is possibility of overfitting. Another challenge that is faced comes from the concern of privacy. Often, sharing of thermal comfort data that are associated with users in a building face the challenge of privacy issues.

To deal with the above challenges, we propose to generate conditional synthetic data for personal thermal comfort models. We propose to use the conditional generative models proposed in ~\citet{das2021conditional} to generate synthetic data for the \textquotedblleft Prefer No Change\textquotedblright, \textquotedblleft Prefer Warmer\textquotedblright and \textquotedblleft Prefer Cooler\textquotedblright classes. The inputs to the generative model are thermal comfort features including physiological signals, temperature, humidity, clothing, activity levels, external parameters etc. The model is capable of extracting the feature representations corresponding to the individual classes, and also to generate new synthetic data keeping the conditional feature representation intact and changing the local noise. Our results show that the proposed model is able to generate synthetic data that mimic the real data.
\section{Related Works}
Synthetic data generation has been proposed to expand the diversity and amount of the existing training data in many different fields, often to improve the robustness of machine learning models. A few examples are as following. In healthcare, \citet{ghorbani2019dermgan} propose a generative adversarial network (GAN ~\citep{goodfellow2014generative,zou2019consensus})-based synthetic data generator to improve the diversity and the amount of skin lesion images. \citet{kohlberger2019whole} synthesize pathology images for cancer with realistic out-of-focus characteristics to evaluate general pathology images for focus quality issues. \citet{Han841619} propose synthetic generation to produce high-resolution artificial radiographs. For privacy-preserving data sharing, \citet{xu2019modeling} propose a method to model tabular data to enable their synthetic generation. In computer vision ~\citet{das2021cdcgen} propose synthetic data generation across multiple domains. In smart buildings, \citet{quintana2020balancing} used a conditional tabular GAN based model for thermal comfort synthetic data generation. We use a state-of-the-art conditional synthetic data generation model that has shown improved results over all baselines to generate thermal comfort synthetic data.
\section{Methods}
\subsection{Thermal Preference Classifier}
Our model is based on the method proposed in ~\citep{das2021conditional}. Suppose we have $N$ samples $\mathbf{X}$ with labels $Y$, with 3 possible thermal preference classes, Warmer/No Change/Cooler. We first train a classifier $C$ (consisting of a feature extractor network denoted by $g(\cdot)$, and a final fully-connected and softmax layer, denoted by $h(\cdot)$, i.e. $C(x) = h(g(x))$) to classify the input sample (which in our case are thermal comfort features) and associated labels as Warmer/No Change/Cooler. Mathematically, this step solves the following minimization with backpropagation:
\begin{align}
    \min_{C} \mathcal{L}_{C}(\mathbf{X},Y) = -\mathbb{E}_{(x,y)\sim (\mathbf{X},Y)}\sum_{l=1}^{2}\left[\mathbb{I}_{[l=y]}\log C(x))\right]
\end{align}

By virtue of the training process, the classifier learns to discard local information and preserve the features necessary for classification (conditional information) towards the downstream layers. Once the classifier is trained, we freeze its parameters, and use it to extract the conditional (Warmer/No Change/Cooler) feature representation $z = g(x)$ (as a vector without spatial characteristics) at the output of the feature extractor network for input image $x$. The dimension of $z$ is chosen such that $\dim(z)$ $<<$ $\dim(x)$.

\subsection{Conditional Generative Flow}\label{meth:cond_gen}
\begin{figure*}[t]
    \centering
    \includegraphics[width=0.85\textwidth]{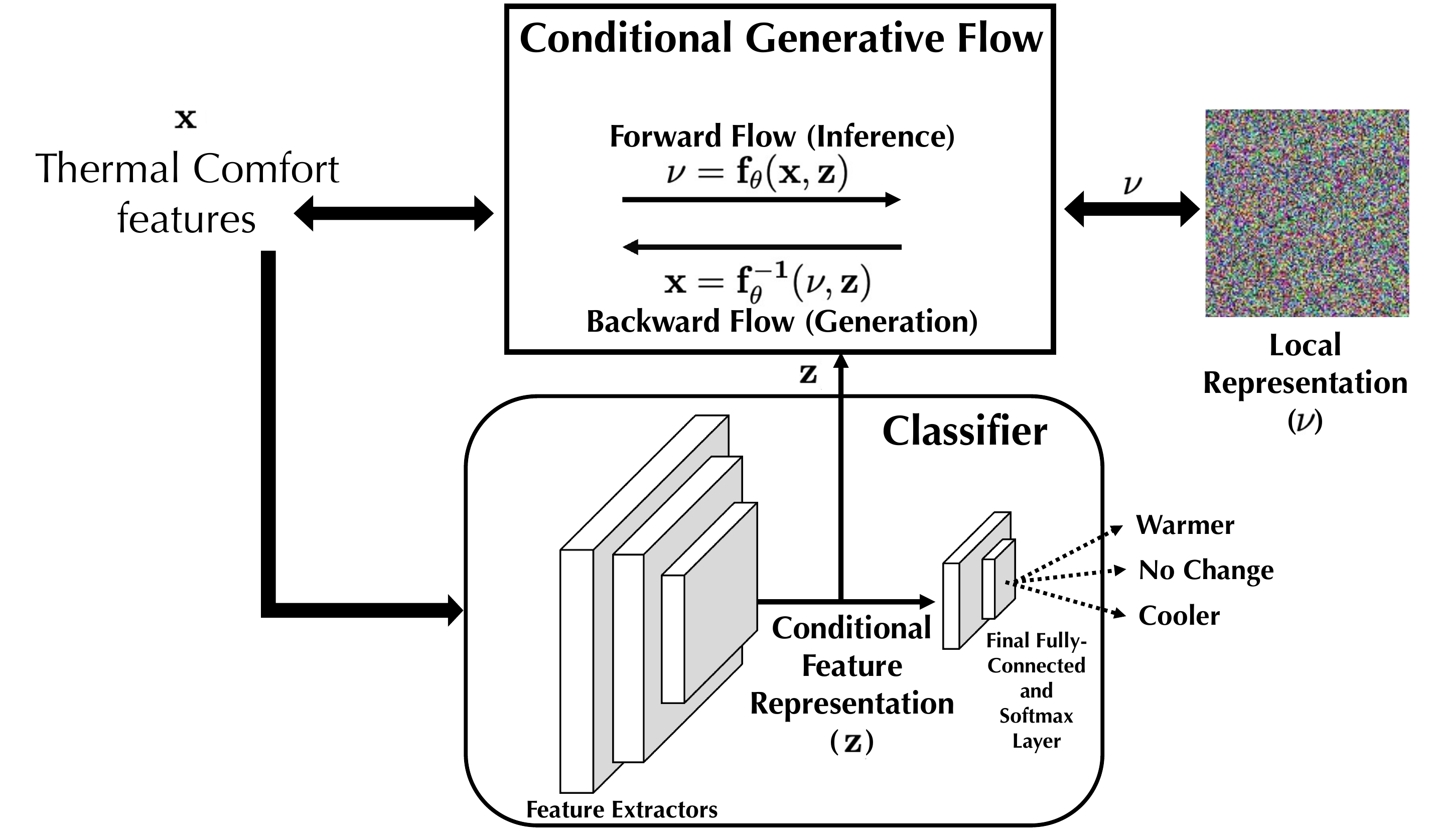}
    \caption{Illustration of the proposed conditional synthetic generation. (Best viewed in color)}
    \label{fig:schematic}
\end{figure*}

During the training phase for the flow model, the conditional feature representation $z$ is fed to the conditional generative flow. The flow model is trained using maximum-likelihood, transforming $x$ to its local representation $\nu$, i.e. 
\begin{align}
    f_{\theta}(x,z) = \nu\sim\mathcal{N}(0,I)
\end{align}
with $\nu$ having the same dimension as $x$ by the inherent design of flow models. We use the method introduced by \citet{das2021conditional,ma2021decoupling} to incorporate the conditional input $z$ in flow model. Coupling layers in affine flow models have scale $(s(\cdot))$ and shift $(b(\cdot))$ networks~\citep{dinh2017density,das2019dimensionality}, which are fed with inputs after splitting, and their outputs are concatenated before passing on to the next layer. We incorporate the conditional information $z$ in the scale and shift networks. Mathematically, (with $x$ as the input, $D$ as input dimension, $d$ as the split size,and $y$ as output of the layer),
\begin{align*}
    x_{1:d},x_{d+1:D} &= \text{split}(x)\\
    y_{1:d} &= x_{1:d}\\
    y_{d+1:D} &= s(x_{1:d},z)\odot x_{d+1:D} + b(x_{1:d},z)\\
    y &= \text{concat}(y_{1:d},y_{d+1:D})
\end{align*}
Since flow models are bijective mappings, the exact $x$ can be reconstructed by the inverse flow with $z$ and $\nu$ as inputs. During the generation phase, for an input sample $x$, we compute the conditional feature representation $z$. Keeping the conditional feature representation the same, we sample a new local representation $\Tilde{\nu}$, and generate a conditional synthetic sample $\Tilde{x}$, i.e.

\begin{align}
    \Tilde{\nu}\in\mathcal{N}(0,I), \;\;\Tilde{x} = f_{\theta}^{-1}(\Tilde{\nu},z)
\end{align}
Here, $\Tilde{x}$ has the same conditional (Warmer/No Change/Cooler) features as $x$ , but has a different local representation. An illustration of the proposed model is provided in Fig.~\ref{fig:schematic}.
\section{Experiments and Results}
In our previous work~\cite{liu2019personal}, we conducted an experiment to collect physiological signals (e.g., skin temperature at various parts of the body, heart rate) of 14 subjects (6 female and 8 male adults) and environmental parameters (e.g., air temperature, relative humidity) for 2–4 weeks (at least 20 h per day). The subjects also took an online survey, where they reported their thermal sensation (on a scale of -3 to +3) and thermal preference (Warmer, Cooler, No Change) among other parameters.

For this work, we generated synthetic data for the 3 thermal preference classes (Warmer, No Change, Cooler) for 5 of the subjects. We designed fully-connected neural networks for the feature extractor, classifier, and conditional generator blocks. A test set is held out from the real dataset to be used for quantitative testing. We then compare the classification performance on this test set for a classifier trained on real data vs a classifier trained on the generated synthetic data. Since the datasets are imbalanced, we report the cohens kappa, accuracy and AUC score (together referred to as classification metrics). 

\begin{table}[t]
\caption{Thermal Preference classification performance with classifiers trained on real and synthetic data. The first number among the pair in each box is performance with a classifier trained on real data, while the second number is with a classifier trained on synthetic data generated by our proposed model.}
\label{tab:results}
\resizebox{1\textwidth}{!}{
\begin{tabular}{p{1cm}|p{2cm}|p{2cm}|p{2cm}|p{2cm}|p{2cm}|p{2cm}}
\toprule
\multicolumn{2}{c}{\multirow{2}{*}{}}                                                                      & \multicolumn{5}{c}{\textbf{Subject ID}}                                                                                                                                    \\ \cline{3-7} 
\multicolumn{2}{c}{}                                                                                       & \multicolumn{1}{c|}{1}              & \multicolumn{1}{c|}{2}               & \multicolumn{1}{c|}{3}               & \multicolumn{1}{c|}{4}               & 5               \\ \hline
\multicolumn{1}{c|}{\multirow{3}{*}{\textbf{Classification Metrics}}} & \multicolumn{1}{c|}{Cohen's Kappa} & \multicolumn{1}{c|}{28.77\%/27\%}   & \multicolumn{1}{c|}{24.59\%/23.12\%} & \multicolumn{1}{c|}{19.23\%/17.91\%} & \multicolumn{1}{c|}{33.65\%/31.78\%} & 18.37\%/15.49\% \\ \cline{2-7} 
\multicolumn{1}{c|}{}                                                 & \multicolumn{1}{c|}{Accuracy}      & \multicolumn{1}{c|}{84.3\%/79.56\%} & \multicolumn{1}{c|}{79.22\%/75.76\%} & \multicolumn{1}{c|}{63.47\%/59.03\%} & \multicolumn{1}{c|}{77.19\%/77.01\%} & 63.22\%/61.42\% \\ \cline{2-7} 
\multicolumn{1}{c|}{}                                                 & \multicolumn{1}{c|}{AUC}           & \multicolumn{1}{c|}{0.81/0.79}      & \multicolumn{1}{c|}{0.8/0.77}        & \multicolumn{1}{c|}{0.67/0.62}       & \multicolumn{1}{c|}{0.78/0.77}       & 0.76/0.74       \\\bottomrule
\end{tabular}}
\end{table}

The classification results for a classifier trained on the real data vs a classifier trained on purely conditional synthetic data, and tested on a hold-out set of real data, is given in Table~\ref{tab:results}. The classifier trained with synthetic data from our proposed model has the close classification performance to that of the classifier trained on real data. This shows the capability of our method to generate synthetic samples with a distribution that closely matches the real conditional data distribution.
\section{Conclusion and Future Work}
We presented preliminary results for thermal comfort synthetic data generation using a state-of-the-art conditional synthetic data generation model. The results show that the generative model is capable of generating synthetic data that are close in distribution with the real data. There are numerous future work to the preliminary work that we have presented. The network of the models can be improved (with e.g. ResNets) for better results. Various scenarios can be explored such as mixing and interpolation in the latent space to generate unseen data. A similar methodology can be extended for synthetic data generation in several more smart building use cases ~\citep{zou2019wifi,zou2019machine,konstantakopoulos2019design,chen2021enforcing,periyakoil2021environmental,das2019novel,das2020occupants,liu2018personal,liu2019personal,donti2021machine,jin2018biscuit}.

\bibliographystyle{abbrvnat}
\bibliography{references}

\end{document}